\pdfoutput=1

\documentclass[11pt]{article}

\usepackage{EMNLP2023}

\usepackage{times}
\usepackage{latexsym,bm}
\usepackage{multirow}
\usepackage{booktabs, adjustbox, amsfonts, amsmath}

\usepackage[T1]{fontenc}

\usepackage[utf8]{inputenc}

\usepackage{microtype}

\usepackage{inconsolata}

\title{Chain-of-Thought Prompt Distillation for Multimodal Named Entity Recognition and Multimodal Relation Extraction}

\author{Feng Chen \\
  Ant Group  \\
  Hangzhou, China \\
  \texttt{chenfeng1271@gmail.com} \\\And
  Yujian Feng \\
  Nanjing University of Posts and Telecommunications \\
  Nanjing, China \\
  \texttt{fengyujian\_904@163.com} \\}

\begin{document}
\maketitle
\begin{abstract}
Multimodal Named Entity Recognition (MNER) and Multimodal Relation Extraction (MRE) necessitate the fundamental reasoning capacity for intricate linguistic and multimodal comprehension. In this study, we explore distilling the reasoning ability of large language models (LLMs) into a more compact student model by generating a \textit{chain of thought} (CoT) -- a sequence of intermediate reasoning steps. Specifically, we commence by exemplifying the elicitation of such reasoning ability from LLMs through CoT prompts covering multi-grain (noun, sentence, multimodality) and data-augmentation (style, entity, image) dimensions.  Subsequently, we present a novel conditional prompt distillation method to assimilate the commonsense reasoning ability from LLMs, thereby enhancing the utility of the student model in addressing text-only inputs without the requisite addition of image and CoT knowledge. Extensive experiments reveal that our approach attains state-of-the-art accuracy and manifests a plethora of advantages concerning interpretability, data efficiency, and cross-domain generalization on MNER and MRE datasets.
\end{abstract}

\section{Introduction}

\begin{figure}[t]
\includegraphics[width=1\linewidth]{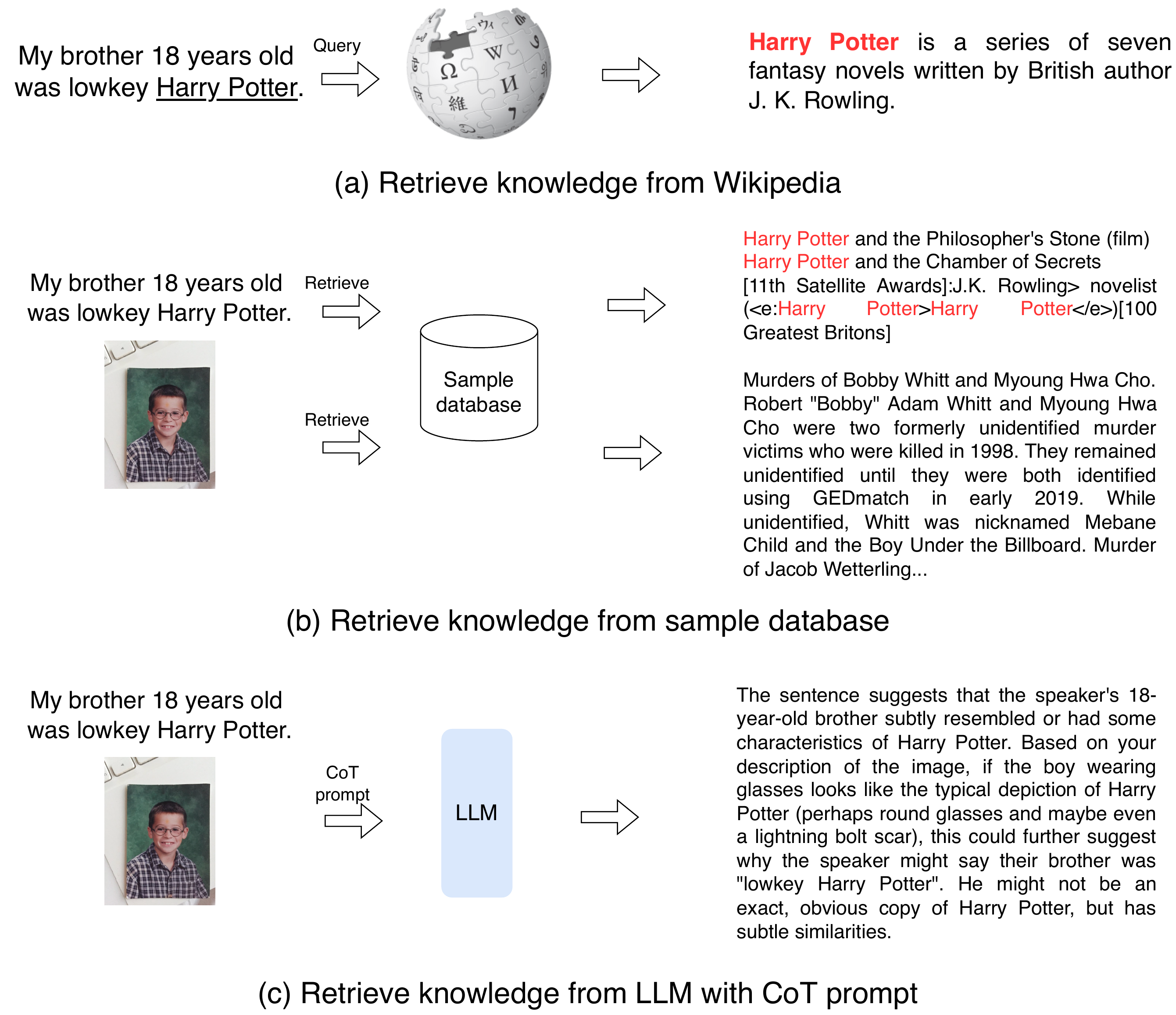}
\caption{Illustration of retrieving knowledge from Wikipedia, sample database and LLMs.}
\label{illustration}
\end{figure}
Multimodal named entity recognition (MNER) \cite{cat,fmit} and multimodal relation extraction (MRE) \cite{hybrid,Zheng2021MultimodalRE} aim to use auxiliary visual clues from images to improve the recognition of multisense or out-of-vocabulary words/relations. However, text-image pairs, amassed from diverse real-world domains such as social media, film critiques, and news articles, pose challenges in comprehending linguistic context and multimodal relationships. A majority of prior methodologies \cite{wang2021improving,gcn,more} employ retrieval augmentation from Wikipedia or sample databases to seek pertinent knowledge concerning images and text, thereby facilitating model reasoning. For instance, KB-NER \cite{wang2021improving} builds a multilingual knowledge base based on Wikipedia to provide related context to the NER model. MoRe \cite{more} utilizes KNN with ViT-B/32 from CLIP \cite{clip} to retrieve the top-k related images from Wikipedia. Consequently, the knowledge procured from these methods exhibits inconsistencies with the current domain \cite{kai2}. For example, the definition of `Harry Potter' in Figure \ref{illustration} (a) is misleading, and the retrieved images in (b) tend to be irrelevant for understanding the query sample. Additionally, these approaches are indirect for interpreting image-text pairs, given that the retrieved knowledge might exhibit semantic discrepancies from the query in the task at hand. As shown in Figure \ref{illustration} (b), the retrieved text from the sample database is hard for understanding the original example. Finally, few of the existing methods provide auxiliary multimodal knowledge that jointly explains the text and image, which is important for MNER and MRE.

Recently, large-scale models have demonstrated remarkable performance on intricate tasks that simulate the reasoning process one might employ when addressing a problem \cite{tot}. Thus, this study investigates the distillation of the reasoning capabilities of large language models (LLMs) into a small model through \underline{C}hain-\underline{o}f-\underline{T}hought \underline{P}rompt \underline{D}istillation (CoTPD). Previous researches \cite{LMfew,Kojima2022LargeLM,Wei2022ChainOT} reveal that the \textit{chain of thought} (CoT) approach favorably elicits multi-step reasoning abilities from LLMs. By generating intermediate natural language rationales that culminate in the final response via CoT prompting, these \textit{chain of thought} demonstrations enable models to delineate a reasoning pathway that deconstructs intricate reasoning into several simpler steps. In this paper, our objective is to harness CoT prompts to synthesize explicit and direct CoT knowledge to understand each sample, and subsequently distill the reasoning capabilities of LLMs into the student model by using such CoT knowledge.  

\textbf{How is CoT knowledge synthesized?} We propose prompting LLMs to interpret each sample in multi-grain, and data-augmentation perspectives. We term the combination of demonstrations with respect to different CoT prompts as CoT knowledge. Multi-grain CoT knowledge includes noun, sentence, and multimodality perspectives. The noun perspective inquires about the definitions of potential entities and specialized vocabulary in the text. The sentence perspective elucidates ambiguous semantics in the original text and supplies necessary background information. The multimodality perspective enables LLMs to interpret the correlation between images and text, explicitly and jointly determining whether an image is helpful for understanding the text.

In addition to multi-grain reasoning, our method inherits the zero-shot reasoning abilities of LLMs through data augmentation. Existing text-based NER methods \cite{aug1,aug2} typically employ rule-based data augmentation to address low-resource and cross-domain challenges. However, the augmented sample is usually homogeneous and unrealistic. In this study, we extend this strategy to the multimodal case through fact-based entity, style, and image augmentation. LLMs are encouraged to perform augmentation according to common sense, enabling the generation of augmented samples that are both diverse and realistic. To the best of our knowledge, our research is the first attempt to utilize multimodal data augmentation in MNER and MRE.

\textbf{How to distill reasoning ability using CoT knowledge?} We believe, combining text-image pair with CoT knowledge as knowledge-enhanced input, i.e., [TXT] + [IMG] + [Knowledge], a small model can easier recognize the entity and relation. As shown in Figure \ref{illustration} (c), CoT knowledge can exceptionally infer the vague relation between `Harry Potter' and the boy. However, such reasoning ability of LLM is not inherited by the small model, since it just simply summarizes the extensive analysis to the final answer, rather than figuring everything out by itself. To solve this issue, we propose a conditional prompt distillation. Specifically, we leverage prompt learning to hint the student model by combining text and a learnable conditional prompt as input, i.e., [TXT] + [Prompt]. By aligning the output distributions predicted from the knowledge-enhanced input view and the prompt-enhanced input view, such contextual CoT knowledge is expected to be distilled in the parameterized prompt. It allows our method to be more practical in dealing with text-only inputs.

The contributions of our method can be summarized in three aspects:

1 We propose a simple but effective method to distill the reasoning ability of LLM to student model, which largely improves the performance with trivial additional cost.


2 Additionally, we propose a fact-based multimodal data augmentation strategy, which exhibits effectiveness in low-resource and cross-domain settings.

3 Our method substantially surpasses existing state-of-the-art approaches with significant improvement in MNER and MRE.




\begin{figure*}[!ht]
\includegraphics[width=1\linewidth]{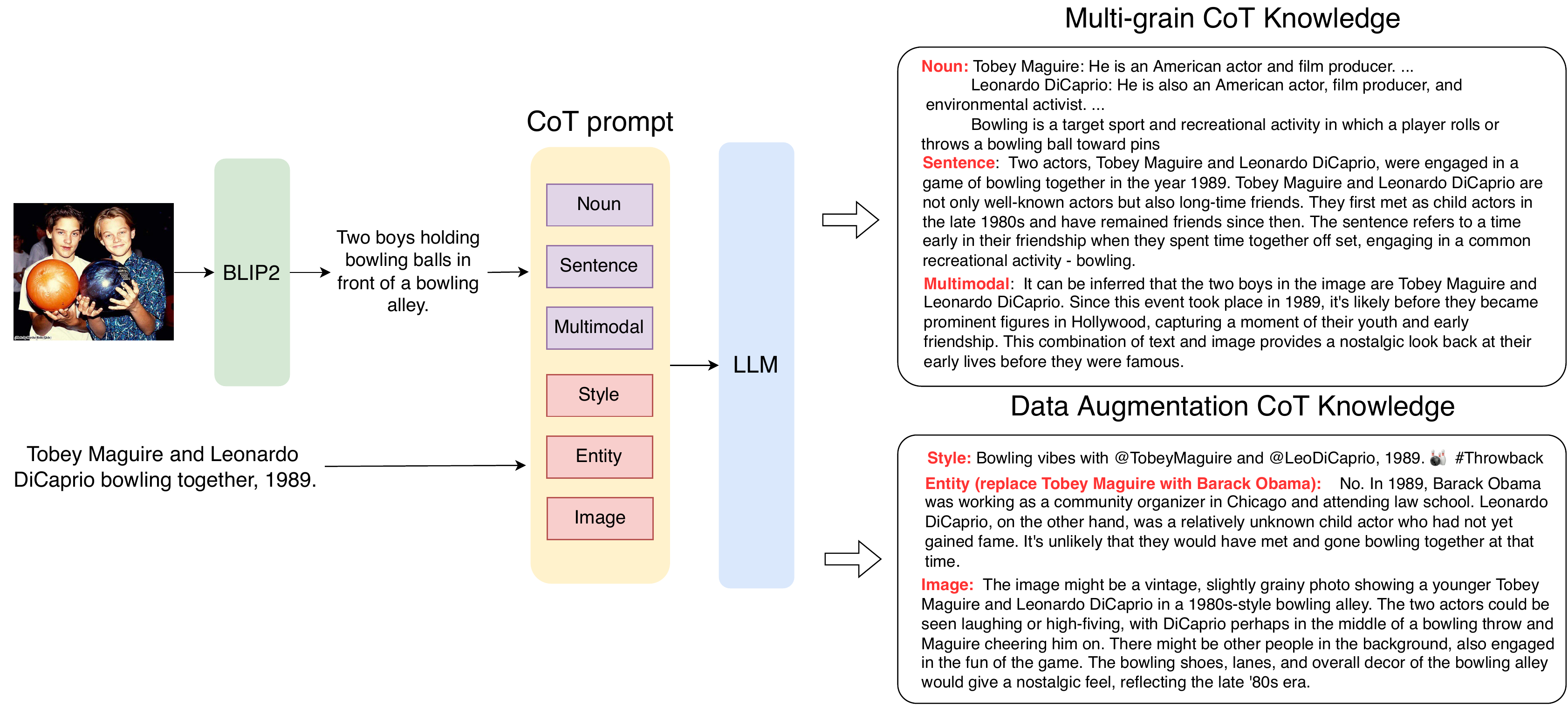}
\caption{Overview of extracting multi-grain CoT knowledge and data augmentation CoT knowledge via querying LLMs with noun, sentence,  multimodality, style, entity, and image CoT prompts.}
\label{cot}
\end{figure*}


\section{Related Work}

\textbf{Chain-of-Thought:} Recently, \textit{chain of thought} (CoT) has been widely used to elicit the multi-step reasoning ability of LLMs, which encourages the LLMs to generate intermediate reasoning chains for solving a problem. \cite{Wei2022ChainOT} prompts LLMs with `Let's think step by step' to facilitate arithmetic, commonsense, and symbolic reasoning. Tree-of-Thoughts \cite{tot} explores performing deliberate decision-making by considering multiple different reasoning paths and self-evaluating choices. In this paper, we propose to use CoT to interpret each sample where the response from LLM  is used as reliable knowledge to help the model understand.

\textbf{Knowledge-based NER and RE} Existing NER and RE methods tend to exploit additional knowledge to assist model reasoning. For example, \cite{wang2021improving} obtains external contexts of a sentence by retrieving and selecting a set of semantically relevant texts through a search engine. Later, Wang et.al \cite{more} extend it to the visual case that MoRe utilizes KNN with ViT-B/32 from CLIP to retrieve the top-k related images from Wikipedia. However, the retrieved knowledge is still vague or misleading for a model to understand. Thus, we propose to use LLMs as  knowledge providers which can accurately explain every detail of each sample.

\section{Our Method}

\subsection{Motivation}

Large language models (LLMs), with their capacity to interpret and process human language in a manner akin to human cognition, facilitate the relation understanding of complex tasks. However, the enormous size of LLMs renders them impractical for industrial use. In the present study, we endeavor to explore an efficient methodology to distill the reasoning prowess of LLMs into a compact model that retains superior performance while ensuring minimal inference latency.

However, traditional knowledge distillation methods \cite{kd} are incompatible to achieve this goal. The reasons are mainly twofold. (1) Prevailing LLMs are predominantly generative models that either produce unstructured output logits \cite{llama} or are solely accessible through official API \footnote{https://openai.com/blog/chatgpt}. (2) Existing LLMs exhibit suboptimal performance in  NER and RE tasks \cite{li2023evaluating, kai1}, attributable to their lack of training under task-specific supervision. To address these issues, we introduce \underline{C}hain-\underline{o}f-\underline{T}hought \underline{P}rompt \underline{D}istillation (CoTPD), a novel approach designed to distill the explicit CoT knowledge into the student model. This method seeks to facilitate knowledge transfer from the parameterized LLM $\stackrel{prompt}{{\longrightarrow}}$ contextual demonstration $\stackrel{distillation}{{\longrightarrow}}$ parameterized student model. Moreover, by training on NER and RE tasks, the student model can obtain promising performance with the help of CoT knowledge.

In addition, existing methods \cite{cat, maf, flat} typically demonstrate their intuition and superiority through case studies or Class Activation Mapping (CAM) visualization. However, these strategies still fall short in providing comprehensive interpretability about how models interpret the multimodal context to reach the final prediction \cite{kai3}. In this investigation, we demonstrate that CoT knowledge, which is akin to how humans understand each sample, benefits the model performance in various domains and tasks.

\subsection{Overview of Our Method}

Given an text and image pair $(\bm{x}, \bm{I})$ where $\bm{x}= \{x_1, x_2, .., x_n\}$ has $n$ words, our method aspires to predict the outputs $\bm{y}$, which could represent the label sequence or the relations for NER and RE tasks respectively. As shown in Figure \ref{cot}, we initially employ image caption technology, i.e., BLIP2 \cite{blip2}, to convert image $\bm{I}$ into an image caption. Subsequently, we utilize langchain\footnote{https://github.com/hwchase17/langchain} with preordained CoT prompts to query the LLM, thereby acquiring CoT knowledge $\bm{c}$ pertaining to each sample. Ultimately, the student language model amalgamates the text-image pair with CoT knowledge as the knowledge-enhanced view, and concatenates the text with a learnable conditional prompt to create a prompt-enhanced view. The objective is to minimize the Kullback-Leibler divergence over the probability distributions of the two views, thereby distilling the contextual CoT knowledge into the parameterized prompt.

\subsection{CoT Prompts for LLMs}
\textbf{Multi-grain CoT Prompts.} To fully understand the text-image pair, we design CoT prompts in noun, sentence, and multimodality perspectives to query LLMs\footnote{Weaker LLM may need accurate descriptions in prompts. We take GPT4 in the Twitter dataset as an example.}, as shown in Figure \ref{cot}.

\textit{Noun:} We query LLMs with `Help me explain the meaning of special words for understanding. + $\bm{x}$', which can  explain potential entities, slang, and terminology.

\textit{Sentence:} For each sample, we query LLMs with `Explain the sentence to me with necessary background. + $\bm{x}$' . It can explain the sentiment, cause, and subject of users.

\textit{Multimodality:} Similarly, we obtain multimodal relation by prompt LLMs with `What is the relation between the text and the attached image? + $\bm{x+I}$'. In this way, LLM can illustrate the vague relation between entity and object and explicitly determine whether the visual information is noise.

\textbf{Data Augmentation CoT Prompts.} We also leverage LLMs to do fact-based  multimodal data augmentation, involving linguistic style, entity, and image perspectives.

\textit{Style:} We query LLMs with `Transform the sentence in Twitter style without changing the meaning. + $\bm{x}$' It diversifies the textual expression with consistent entity and meaning.

\textit{Entity:} Initially, we adhere to \cite{metaaug4ner} to substitute the entity in $\bm{x}$ with an entity of the same type sourced from the MNER dataset. For instance, we replace the entity 'Tobey Maguire [PER]' with 'Barack Obama [PER]'. Then, we assess the factual accuracy of the augmented sample by prompting the LLM with the query 'Whether the sentence is possible in fact, answer yes or no. + $\bm{x}$'. We retain only the samples that receive a positive response indicating factual plausibility.

\textit{Image:} We generate the visual information of text by querying LLMs `What is a possible image with the text in a tweet? + $\bm{x}$'

We unify multi-grain CoT knowledge with the original image and text pair. For data augmentation CoT knowledge, we reorganize it as a new sample. Notably, the total knowledge may exceed the max token limits for the student model, therefore, we first summarize the over-length knowledge.

\subsection{Conditional Prompt Distillation}

To distill CoT knowledge $\bm{c}$ into a compact model, we first try multi-view alignment \cite{ita} to match the txt-only output distribution and text-image-CoT output distribution. However, we find it results in suboptimal performance (Sec \ref{sec ablation}). We believe the reasons are twofold: (1) CoT knowledge is higher-level context than previous retrieve knowledge. (2) The knowledge is personalized which needs extra parameters to distill. Therefore, inspired by prompt learning \cite{prefix,cocoop}, we propose a conditional prompt distillation with respect to the text to hint the student model adjustably.

We introduce our conditional prompt generator, which is composed of a transformer decoder with learnable queries $\bm{q} \in \mathbb{R}^{N \times d}$. The conditional prompt $\bm{p}$ is generated by:

\begin{align}
 \bm{p}=Softmax(\frac{(\bm{q} W_q)(\bm{x} W_k)^T}{\sqrt{d}}) \cdot \bm{x} W_v,
\end{align}
where $W_q, W_k$, and $W_v$ are learnable projectors for query, key, and value. We concatenate text, image caption, and CoT knowledge as the knowledge-enhanced view $[\bm{x}; \bm{I}; \bm{c}]$. Similarly, the prompt-enhanced view is composed by concatenating text and conditional prompt as $[\bm{x}; \bm{p}]$. We feed these two views to the same text encoder $E$ separately:

\begin{equation}
\begin{split}
   H_{k} =\{h_1,..,h_n,..,h_{n+l}\} = E([\bm{x}; \bm{I}; \bm{c}]), \\
    H_{p} = \{h_1,..,h_n,..,h_{n+l}\} = E([\bm{x}; \bm{p}]),
\end{split}
\end{equation}
where $H = \{h\}_1^{n+l}$ denotes the output of encoder with max token limit $n+l$. The subscripts $k$ and $t$ represent knowledge-enahcned and prompt-enhanced view respectively. The former $n$ tokens correspond to the embedding of the text part in two views, which are used for following NER or RE prediction.

Then, we reduce the gap between these two views over the output distributions, so that the CoT knowledge and image are distilled into the student model under specific task. We denote it as conditional prompt distillation loss:
\begin{equation}
\begin{split}
    \mathcal{L}_{CPD} &=KL\left(p_\theta(\boldsymbol{y} \mid \bm{x}, \bm{I}, \bm{c}) \| p_\theta(\boldsymbol{y} \mid \bm{x}, \bm{p})\right) \\
    &= KL(H_k[1:n] || H_t[1:n])
\end{split}
\end{equation}

Besides, for MNER and MRE, we use the negative log-likelihood (NLL) as the training loss for the  knowledge-enhanced view with gold labels $\bm{y}^*$:

\begin{equation}
    \mathcal{L}_{\mathrm{NLL}}=-\log p_{\theta}\left(\boldsymbol{y}^* \mid \boldsymbol{x}, \boldsymbol{I}, \boldsymbol{c}\right)
\end{equation}

Thus, the total loss $\mathcal{L}_{total}$ is the combination of $\mathcal{L}_{CPD}$ and $\mathcal{L}_{NLL}$ with coefficient factor $\alpha$:

\begin{equation}
    \mathcal{L}_{total} =  \mathcal{L}_{NLL} + \alpha \cdot \mathcal{L}_{\mathrm{CPD}}
\end{equation}

\section{Experiments}
\subsection{Settings}

\textbf{Datasets.} For MNER evaluation, we verify the effectiveness of our method on Twitter2015 \cite{twitter15}, Twitter2017 \cite{twitter17}, SNAP \cite{snap} and WikiDiverse \cite{more}. These datasets containing 4,000/1,000/3,357, 3,373/723/723 and 4,290/1,432/1,459, 6,312/755/757 sentences in train/development/test split respectively. The former three datasets are collected from social media domain, while WikiDiverse is reorganized by \cite{more} from Wikinews as a multi-modal NER dataset. For MRE, we use MNRE dataset \cite{mnre} constructed by user tweets, which contains 12,247/1,624/1,614 sentences in train/development/test split.

\textbf{Model Configuration.} To fairly compare with
most of the recent works, we adopt BERT-base-uncased as our text encoder. Besides, we follow \cite{ita, more} to adopt XLM-RoBERTa-large (XLMR) to achieve state-of-the-art performance. For LLM prompting, we use langchain to query each sample in a dialogue way and record the answer of LLMs as CoT knowledge. The version of ChatGPT used in our experiments is gpt-3.5-turbo and the version of GPT4 is gpt-4.  All LLMs used in our method set the sampling temperature to 0 for stable output.

\textbf{Training Configuration.} Our approach is implemented in Pytorch framework on an NVIDIA A100 GPU. We adopt AdamW optimizer to minimize the loss function. Besides, we follow \cite{more} to grid search the learning rate of our models within [$1\times10^{-6}$, $5\times10^{-5}$]. The max length of sentence input is set to 512 to cover more CoT knowledge. The model undergoes a fixed 15 epochs of training with 32 mini-batch.

\textbf{Baseilne Model and Variants.} We refer to a model training and testing with [TXT]+[IMG] input as our baseline model. For the knowledge variants, we take retrieve knowledge from MoRe \cite{more} as Retri, and  knowledge queried from Wikipedia \cite{wang2021improving} as Wiki, and our CoT knowledge as CoT. For Retri, MoRe provides image-retrieved knowledge Retri$_{txt}$ and text-retrieved knowledge Retri$_{img}$. For distillation variants, we refer to multi-view alignment from ITA \cite{ita} as MV. Furthermore, in our method, considering different kinds of learnable prompts, we compare prefix \cite{prefix} and unconditional prompt \cite{coop}, which are denoted as PrefixD and UPD respectively, to verify the effectiveness of the proposed conditional prompt distillation (CPD). 


\subsection{Main Result}
 We compare our approach with other BERT-based \cite{cat,fmit,gcn} and XLMR-based \cite{ita,promptmner,more,cat} state-of-the-art methods on MNER and MRE. To fully show the effectiveness of our approach, we evaluate ChatGPT and GPT4 on Twitter2015, Twitter2017 and MNRE. For our approach, we also show the model performance using different LLM teacher and student models. As shown in Table \ref{tab main}, we observe LLMs are inferior to existing state-of-the-art methods, but the CoT knowledge significantly boosts our method. We think the training data of LLMs includes the dataset but the training of LLM is not in NER or RE supervision. Therefore, LLMs can generally understand each sample but still underperform in specific tasks. Besides, compared with other previous SOTA methods, our approach suppresses them by a large margin. Compared to MoRe, our method achieves 0.82\% and 5.96\% improvement on Twitter2015 and MNRE respectively. We attribute such significant gain to the direct and comprehensive CoT knowledge from LLM, so that the student model can easily understand the multimodal sample, instead of analogically learning from the related samples.  We also observe that the ability of LLMs and student models significantly influences the final performance. For example, with respect to the same student XMLR model, GPT4 can generally provide a further 0.4\%-1.2\% improvement over ChatGPT. Similarly, a powerful student model also brings large performance boosts.

\begin{table}[t]
\begin{adjustbox}{width=1\columnwidth}
\begin{tabular}{c|c|ccccc}
\toprule
                      & Method            & T-15 & T-17 & SNAP & Wiki & MNRE \\ \midrule
\multirow{3}{*}{BERT} & CAT-MNER          &    75.41  &    85.99  &  -    &   -   &  -    \\
                      & R-GCN             &  75.00    &  87.11    &   -   &   -   &   -   \\
                      & FMIT              &   76.25   &  86.79    &  -    &   -   &   -   \\ \midrule
\multirow{4}{*}{XMLR} & ITA               &   78.03   &  89.75    &  90.15    &  76.87    &   66.89   \\
                      & promptMNER        &   78.60   &   90.27   &   -   &  -    &  -    \\
                      & CAT-MNER          & 78.72     &   90.47   &   -   &   -   &  -    \\
                      & MoRe              &  79.21    &  90.67    &  91.10    &  79.33    &  68.60    \\ \midrule
\multirow{2}{*}{LLM}  & ChatGPT           &  50.21    &  57.50    &  -    &   -   &  35.20    \\
                      & GPT4              &  57.98    &  66.61    & -     & -     &  42.11    \\ \midrule
\multirow{4}{*}{}     & our(ChatGPT-BERT) &  76.53    & 87.79     &   88.12   &  76.42    & 66.42     \\
                      & our(GPT4-BERT)    &  76.90    &  88.27    &  88.51    &  77.09    &  67.88    \\
                      & our(ChatGPT-XMLR) &    \textbf{80.31}  & 91.78     & 91.97     &   80.02   &  73.67    \\
                      & our(GPT4-XMLR)    & 80.03     & \textbf{92.20}     & \textbf{92.41}     & \textbf{81.20}     & \textbf{74.56}     \\ \bottomrule
\end{tabular}
\end{adjustbox}
\caption{Performance comparison (F1 score) with existing methods and LLMs. The evaluation of LLM is based on \cite{li2023evaluating} with multimodal modification. For our method, such as `ours(GPT4-XMLR)', the two models in brackets denote the teacher LLM and student model respectively.}\label{tab main}
\end{table}

\begin{table}[]
\begin{adjustbox}{width=0.95\columnwidth}
\begin{tabular}{l|ccccc}
\toprule
                      & T-15                 & T-17                 & SNAP                 & Wiki                 & MNRE                 \\ \midrule
our method            & \multicolumn{1}{l}{80.03} & \multicolumn{1}{l}{92.20} & \multicolumn{1}{l}{92.41} & \multicolumn{1}{l}{81.20} & \multicolumn{1}{l}{74.56} \\
\ \ \  w/o Noun              &   78.63                   & 89.70                     &  90.56                    & 77.32                     &        70.20              \\
\ \ \ w/o Sentence          &  79.04                    &  89.87                    & 91.21                     &  79.09                    &  69.41                    \\
\ \ \ w/o Multimodality        &   79.10                   &   91.20                   &  91.56                    &  79.32                    &    73.21                  \\
\ \ \ w/o Style &  79.65                    &  91.70                    &  91.81                    &    81.04                  &   74.16    \\
\ \ \ w/o Entity &      79.00                &   90.98                   &    92.35                  &  79.56                    &  -     \\

\ \ \ w/o Image & 79.65                     &   91,43                   & 91.40                     &    79.66                  &    73.98   \\
\bottomrule
\end{tabular}
\end{adjustbox}\caption{Effect of multi-grain and data augmentation CoT knowledge. Notably, `w/o Image' means the augmented samples do not have any visual information.}\label{tab cot}
\end{table}

\subsection{Ablation Study} \label{sec ablation}

\textbf{Effectiveness of Different CoT Knowledge.}  Multi-grain CoT knowledge focuses on explaining each sample from different perspectives, leading to significant improvement by enhancing the understanding of the student model. Specifically, as shown in Table \ref{tab cot}, noun  knowledge contributes more to MNER than MRE while sentence knowledge brings more improvement to MRE. We believe these two kinds of knowledge meet the demand of two tasks where MNER focuses on entity mention and MRE concentrates on the entity relation. Besides, it is worth noting that data augmentation CoT knowledge, i.e., style, entity, and image augmentation, contributes less than multi-grain CoT knowledge. It is because data augmentation mainly generates more samples for training, rather than directly explaining each sample. Therefore, it is more effective in cross-domain and low-resource cases, which will be elaborated on next section.

\begin{table}[]
\begin{adjustbox}{width=0.95\columnwidth}
\begin{tabular}{l|ccccc}
\toprule
                     & T-15                 & T-17                 & SNAP                 & Wiki                 & MNRE                 \\
                     \midrule
baseline             & \multicolumn{1}{c}{77.21} & \multicolumn{1}{c}{88.77} & \multicolumn{1}{c}{89.53} & \multicolumn{1}{c}{76.04} & \multicolumn{1}{c}{65.85} \\
baseline+Wiki         & 78.15                     &    89.60                  &   -                   &    -                  &    67.21                  \\
baseline+Retri$_{txt}$         &  78.05                     &   89.87                   &    90.44                  &  78.24                    &  -                    \\
baseline+Retri$_{img}$         &  78.35                     &   90.17                   &    90.02                 &  77.94                    &  67.60                    \\
baseline+CoT         & 80.24                     &  91.92                    &  92.40                    &    81.00                  &  74.74                    \\ \midrule

baseline+Wiki+MV         & 78.26                     &    89.22                  &   -                   &    -                  &    67.01                  \\
baseline+Retri$_{txt}$+MV     &   78.29                   &     90.47                 &  90.31                    &  77.82                    &  -                    \\
baseline+Retri$_{img}$+MV     &   78.20                   &     90.12                 &  90.11                    &  78.03                    &  67.54                    \\

baseline+CoT+MV     &  79.62                    &  91.90                    &  91.98                    &    81.32                  &  70.41                    \\

baseline+CoT+UPD &    79.34                  &   91.02                   &  91.50                     &    80.00                  & 70.45     \\
baseline+CoT+PrefixD &   79.62                   &  91.57                    &          92.20            &  80.62                    &   71.62   \\
baseline+CoT+CPD &    80.03                  &   92.20                   &   92.41                   &    81.20                  &  74.56    \\ \bottomrule
\end{tabular}
\end{adjustbox}\caption{Ablation study of knowledge variants and distillation variants.}\label{knowledge}
\end{table}

\textbf{Comparison with Previous Retrieved Knowledge.} In Table \ref{knowledge}, we also report the results using retrieved knowledge to verify the effectiveness of our CoT knowledge. For retrieved knowledge from Wikipedia and the sample database, our CoT knowledge is better than them by a large margin. For example, CoT suppresses Wiki with 2.09\% gains and outperforms Retri$_{txt}$ with 2.19\% improvements on Twitter2015. It demonstrates that such direct CoT knowledge is more effective to understand each sample. Furthermore, we investigate how knowledge difference influences the final prediction in Appendix. It shows accurate and direct CoT knowledge contains better potential in interpretability and performance over existing methods.

\textbf{Distilling Knowledge into Student Model.} As shown in Table \ref{knowledge}, we also compare the results of different knowledge+distillation variants on MNER and MRE. It is obvious that multi-view alignment is unsuitable for distilling CoT knowledge to the student model, where `baseline+CoT+MV' is 4.33\% lower than `baseline+CoT' on MNRE. We think such explicit and direct knowledge  is higher-level than the retrieved knowledge. Thus, distilling it into the student model is harder. Alternatively, our conditional prompt distillation can maintain the performance with minimal decrease. Moreover, we test other two kinds of prevailing learnable prompts for distillation, i.e., prefix and unconditional prompt. It shows the proposed conditional prompt is better than them. We think the personalized/conditional prompt with respect to the text can hint the student model easier.

\subsection{Detailed Analysis}

\textbf{Cross-domain Generalization.} The difference in type distribution and data characteristics often brings significant performance gaps in practice. Thus, we examine the cross-domain generalization of our method in Table \ref{cross}. First, without data augmentation, our method achieves comparable results as the previous best model where our method suppresses CAT-MNER with 0.68\% F1 improvement but is inferior to it with 0.28\% decrease in two cross-domain settings respectively.

Second, when the entity augmentation uses entities from the source dataset (in-domain aug), our method achieves 1.59\% and 0.73\% F1 gains over CAT-MNER. It demonstrates the data augmentation strategy can substantially boost generalization ability. Besides, in this work, we aim to let the student model inherit the zero-shot ability of LLM. Therefore, we first query LLMs to provide a list of 400 Twitter entities, covering four types but without overlapping the entities in the source set, and then apply the novel entities for entity replacement. We refer to this implementation as zero-shot augmentation. As shown in Table \ref{cross}, augmenting novel entities can further provide a 0.3\%-0.6\% F1 improvement. This demonstrates the effectiveness of utilizing data-augmentation CoT knowledge for achieving cross-domain generalization

\begin{table}[]
\centering
\begin{adjustbox}{width=1.0\columnwidth}
\begin{tabular}{c|ccc|ccc}
\toprule
    \multirow{2}*{Methods} & \multicolumn{3}{c}{Twitter2017$\rightarrow$Twitter2015} & \multicolumn{3}{|c}{Twitter2015$\rightarrow$Twitter2017} \\
    & P & R & F1 & P & R & F1 \\
\midrule
	UMT       & 64.67       & 63.59       & 64.13       & 67.80       & 55.23       & 60.87       \\
	UMGF      & 67.00       & 62.18       & 66.21       & 69.88       & 56.92       & 62.74       \\
	FMIT      & 66.72       & 69.73       & 68.19       & 70.65       & 59.22       & 64.43       \\
	CAT-MNER  & 74.86       & 63.01       & 68.43       & 70.69       & 59.44       & 64.58       \\
	our(w/o aug)      &  75.76      &   64.06    &  69.11      &  70.08      &   59.67      &    64.30    \\
	our(w/ in-domain aug))     &  76.51      &   64.21    &  70.02      &  71.64      &   59.49      &    65.31    \\
 	our (w/ zero-shot aug)     & 76.04      & 66.28      &  70.40      &  71.34      &   60.16      &  66.05      \\
\bottomrule
\end{tabular}
\end{adjustbox}\caption{Comparison of the cross-domain generalization ability. For fairness, we use GPT4 as teacher LLM and BERT$_{base}$ as student backbone. Results of other methods are from \cite{cat,flat}.}
\label{cross}
\end{table}

\begin{figure}[!ht]
\includegraphics[width=0.9\linewidth]{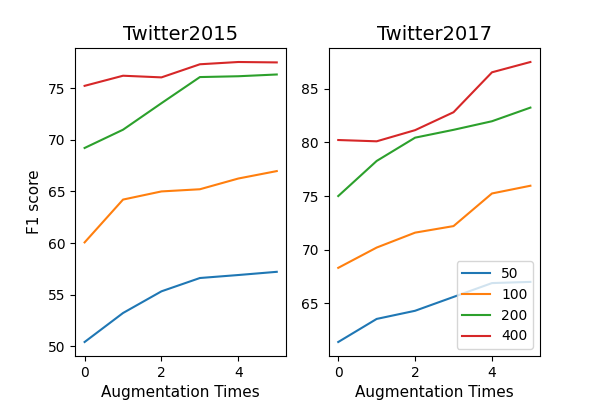}
\caption{Influence of augmentation times with $\{50, 100, 200, 400\}$ training data on Twitter2015 and Twitter2017. Times=0 means using training data without any augmentation.}
\label{data}
\end{figure}

\textbf{Augmentation Times for Data Efficiency.} The size of augmented samples is essential for final model performance.  Typically, we randomly sample $\{50, 100, 200, 400\}$ image-text pairs from Twitter2015 and Twitter2017 with different augmentation times to examine the effectiveness of our multimodal augmentation strategy. As depicted in Figure \ref{data}, increased augmentation yields improved performance, particularly when larger examples are utilized. Notably, our augmentation strategy demonstrates a linear enhancement with the progression of augmentation times. We attribute this improvement to the diversity and authenticity from linguistic style, factual entity replacement and image imagination, rather than simply substituting entities \cite{metaaug4ner}.

\textbf{Influence of BLIP2.} We observe that BLIP2 \cite{blip2} has basic visual-entity alignment ability. It means it may recognize the entity mentions in the image, which may be unfair for performance comparison. We discuss the influence of BLIP2 in Table \ref{blip} and observe that, for both ITA and our method, BLIP2 provides 0.3\%-0.5\% improvement over VinVL \cite{vinvl} on two datasets. However, even using VinVL for image caption, our method can still achieve  significant improvement over the previous best method MoRe. We think LLM can infer visual-entity relation from the general noun description, as shown in the multimodal CoT knowledge in Figure \ref{cot}.

\begin{table}[]
\centering
\begin{tabular}{c|c|cc}
\toprule
Image Caption             & Methods & T-15 & T-17 \\
\midrule
\multirow{2}{*}{VinVL} & ITA     &   78.03   &  89.75    \\
                       & our     & 79.65     &  92.03     \\ \midrule
\multirow{2}{*}{BLIP2} & ITA     &  78.32    &   90.22   \\
                       & our     & 80.03     &  92.20   \\ \bottomrule
\end{tabular}\caption{Influence of using different image caption methods.}\label{blip}
\end{table}

\begin{figure*}[t]
\includegraphics[width=1\linewidth]{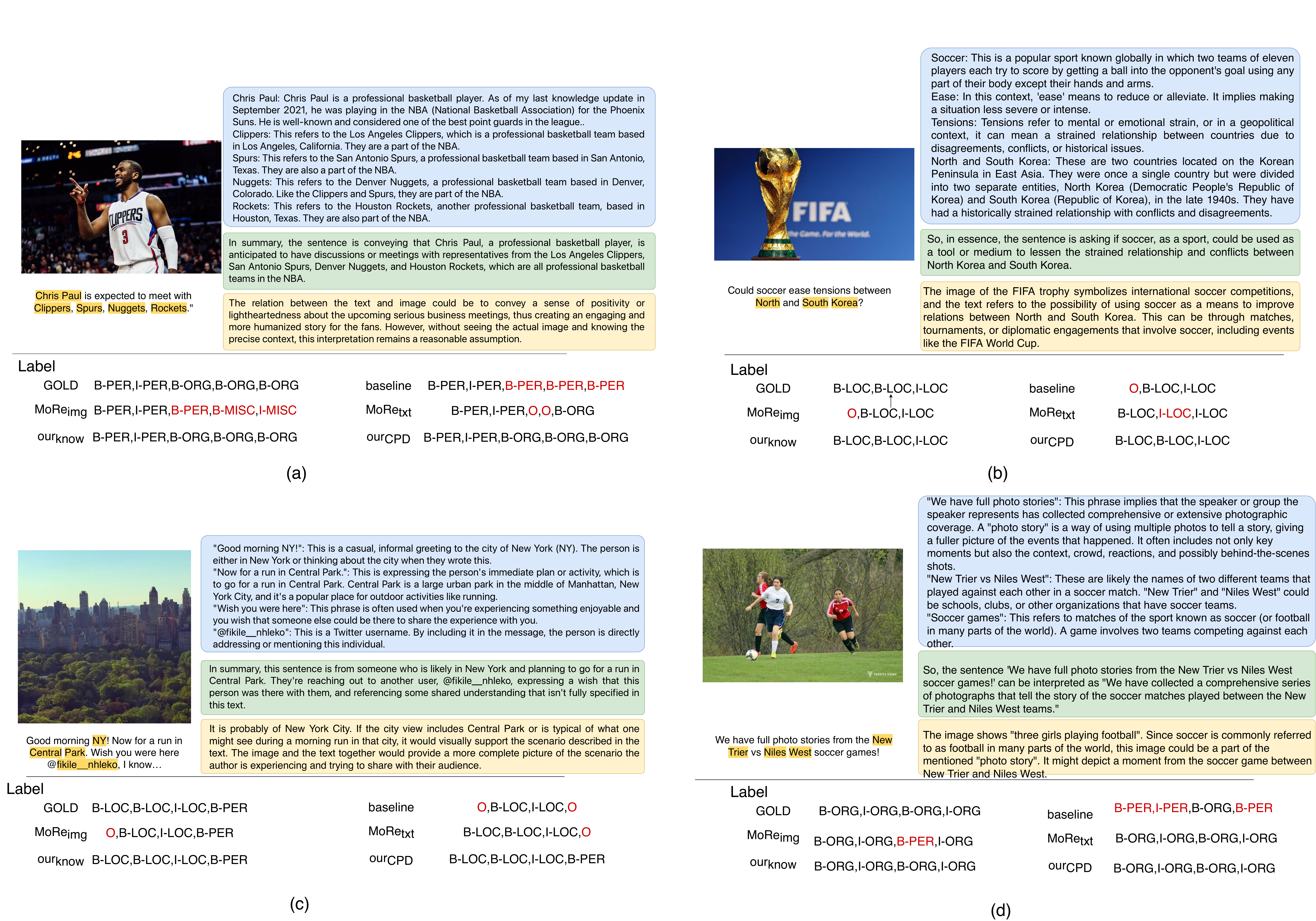}
\caption{Case studies of how CoT knowledge can help model predictions. The subscript $know$ and $CPD$ denote inference with and without CoT knowledge respectively.}
\label{case}
\end{figure*}

\textbf{Case Study.} We showcase how CoT knowledge can help improve the predictive performance of the model. As illustrated in Figure \ref{case}, LLM first provides a detailed explanation for `special' words in the text, largely enriching the linguistic semantics. Especially for abbreviation, slang, and hashtag, LLMs directly return crucial definitions for easy understanding of the whole context while KB-NER and MoRe only provide related or vague context needing further consideration. For multimodal CoT knowledge, our method can distinguish the relatedness of text-image pair. For example, in (b) and (d), the images do not have a corresponding entity in the text. However, LLMs can also interpret the long-range image-text relation.

Furthermore, compared with the baseline model and MoRe, our method can easily recognize hard entities like `North and South Korea' in (b) and `fikile\_nhleko' in (c). Besides, the correct prediction with and without CoT knowledge indicates that the student model inherits the reasoning ability of LLMs through conditional prompt distillation.

\textbf{Limitations and Boarder Impacts.} We notice that the quality of generated CoT knowledge heavily depends on the well-posed LLM that understands the prompt and example comprehensively. We believe next-generation multimodal large models, such as GPT5, will make the CoT knowledge more reliable.

Our method is simple yet effective to distill the general task-agonist knowledge from LLMs to a small model. It can be easily expanded to other tasks which need sophisticated reasoning.

\section{Conclusion}

In this paper, we propose a novel CoT knowledge covering multi-grain understanding of examples and data-augmentation generalization from LLMs. Furthermore, we introduce conditional prompt distillation to distill the reasoning ability of LLMs to a compact student model. Extensive experiments on five datasets show that our method outperforms all existing state-of-the-art methods and manifests a plethora of advantages concerning interpretability, data efficiency, and cross-domain generalization.

\bibliography{anthology,custom}

\begin{thebibliography}{34}
\expandafter\ifx\csname natexlab\endcsname\relax\def\natexlab#1{#1}\fi

\bibitem[{Brown et~al.(2020)Brown, Mann, Ryder, Subbiah, Kaplan, Dhariwal,
  Neelakantan, Shyam, Sastry, Askell, Agarwal, Herbert-Voss, Krueger, Henighan,
  Child, Ramesh, Ziegler, Wu, Winter, Hesse, Chen, Sigler, Litwin, Gray, Chess,
  Clark, Berner, McCandlish, Radford, Sutskever, and Amodei}]{LMfew}
Tom~B. Brown, Benjamin Mann, Nick Ryder, Melanie Subbiah, Jared Kaplan,
  Prafulla Dhariwal, Arvind Neelakantan, Pranav Shyam, Girish Sastry, Amanda
  Askell, Sandhini Agarwal, Ariel Herbert-Voss, Gretchen Krueger, T.~J.
  Henighan, Rewon Child, Aditya Ramesh, Daniel~M. Ziegler, Jeff Wu, Clemens
  Winter, Christopher Hesse, Mark Chen, Eric Sigler, Mateusz Litwin, Scott
  Gray, Benjamin Chess, Jack Clark, Christopher Berner, Sam McCandlish, Alec
  Radford, Ilya Sutskever, and Dario Amodei. 2020.
\newblock Language models are few-shot learners.
\newblock \emph{ArXiv}, abs/2005.14165.

\bibitem[{Chen et~al.(2020)Chen, Wang, Tian, Yang, and Yang}]{aug2}
Jiaao Chen, Zhenghui Wang, Ran Tian, Zichao Yang, and Diyi Yang. 2020.
\newblock Local additivity based data augmentation for semi-supervised ner.
\newblock \emph{arXiv preprint arXiv:2010.01677}.

\bibitem[{Chen et~al.(2021)Chen, Aguilar, Neves, and Solorio}]{aug1}
Shuguang Chen, Gustavo Aguilar, Leonardo Neves, and Thamar Solorio. 2021.
\newblock Data augmentation for cross-domain named entity recognition.
\newblock \emph{arXiv preprint arXiv:2109.01758}.

\bibitem[{Chen et~al.(2022)Chen, Zhang, Li, Deng, Tan, Xu, Huang, Si, and
  Chen}]{hybrid}
Xiang Chen, Ningyu Zhang, Lei Li, Shumin Deng, Chuanqi Tan, Changliang Xu, Fei
  Huang, Luo Si, and Huajun Chen. 2022.
\newblock Hybrid transformer with multi-level fusion for multimodal knowledge
  graph completion.
\newblock \emph{arXiv preprint arXiv:2205.02357}.

\bibitem[{Gou et~al.(2021)Gou, Yu, Maybank, and Tao}]{kd}
Jianping Gou, Baosheng Yu, Stephen~J Maybank, and Dacheng Tao. 2021.
\newblock Knowledge distillation: A survey.
\newblock \emph{IJCV}, 129:1789--1819.

\bibitem[{Kai~Zhang(2021)}]{kai2}
Ruobing Xie Xu Han Zhiyuan Liu Fen Lin Leyu Lin Maosong~Sun Kai~Zhang,
  Yuan~Yao. 2021.
\newblock Open hierarchical relation extraction.
\newblock In \emph{NAACL}.

\bibitem[{Kai~Zhang(2023)}]{kai1}
Yu~Su Kai~Zhang, Bernal Jiménez~Gutiérrez. 2023.
\newblock Aligning instruction tasks unlocks large language models as zero-shot
  relation extractors.
\newblock In \emph{Findings of ACL}.

\bibitem[{Kojima et~al.(2022)Kojima, Gu, Reid, Matsuo, and
  Iwasawa}]{Kojima2022LargeLM}
Takeshi Kojima, Shixiang~Shane Gu, Machel Reid, Yutaka Matsuo, and Yusuke
  Iwasawa. 2022.
\newblock Large language models are zero-shot reasoners.
\newblock \emph{ArXiv}, abs/2205.11916.

\bibitem[{Li et~al.(2023{\natexlab{a}})Li, Fang, Yang, Wang, Ye, Zhao, and
  Zhang}]{li2023evaluating}
Bo~Li, Gexiang Fang, Yang Yang, Quansen Wang, Wei Ye, Wen Zhao, and Shikun
  Zhang. 2023{\natexlab{a}}.
\newblock Evaluating chatgpt's information extraction capabilities: An
  assessment of performance, explainability, calibration, and faithfulness.
\newblock \emph{arXiv preprint arXiv:2304.11633}.

\bibitem[{Li et~al.(2023{\natexlab{b}})Li, Li, Savarese, and Hoi}]{blip2}
Junnan Li, Dongxu Li, Silvio Savarese, and Steven Hoi. 2023{\natexlab{b}}.
\newblock Blip-2: Bootstrapping language-image pre-training with frozen image
  encoders and large language models.
\newblock \emph{arXiv preprint arXiv:2301.12597}.

\bibitem[{Li and Liang(2021)}]{prefix}
Xiang~Lisa Li and Percy Liang. 2021.
\newblock Prefix-tuning: Optimizing continuous prompts for generation.
\newblock \emph{arXiv preprint arXiv:2101.00190}.

\bibitem[{Li et~al.(2020)Li, Yan, Qiu, and Huang}]{flat}
Xiaonan Li, Hang Yan, Xipeng Qiu, and Xuanjing Huang. 2020.
\newblock Flat: Chinese ner using flat-lattice transformer.
\newblock \emph{arXiv preprint arXiv:2004.11795}.

\bibitem[{Lu et~al.(2018)Lu, Neves, Carvalho, Zhang, and Ji}]{snap}
Di~Lu, Leonardo Neves, Vitor Carvalho, Ning Zhang, and Heng Ji. 2018.
\newblock Visual attention model for name tagging in multimodal social media.
\newblock In \emph{ACL}, pages 1990--1999.

\bibitem[{Lu et~al.(2022)Lu, Zhang, and Zhang}]{fmit}
Junyu Lu, Dixiang Zhang, and Pingjian Zhang. 2022.
\newblock Flat multi-modal interaction transformer for named entity
  recognition.
\newblock \emph{arXiv preprint arXiv:2208.11039}.

\bibitem[{Radford et~al.(2021)Radford, Kim, Hallacy, Ramesh, Goh, Agarwal,
  Sastry, Askell, Mishkin, Clark, Krueger, and Sutskever}]{clip}
Alec Radford, Jong~Wook Kim, Chris Hallacy, Aditya Ramesh, Gabriel Goh,
  Sandhini Agarwal, Girish Sastry, Amanda Askell, Pamela Mishkin, Jack Clark,
  Gretchen Krueger, and Ilya Sutskever. 2021.
\newblock Learning transferable visual models from natural language
  supervision.
\newblock In \emph{ICML}, pages 8748--8763.

\bibitem[{Sun et~al.(2019)Sun, Zhang, Ji, and Yang}]{kai3}
Lin Sun, Kai Zhang, Fule Ji, and Zhenhua Yang. 2019.
\newblock {TOI}-{CNN}: a solution of information extraction on {C}hinese
  insurance policy.
\newblock In \emph{NAACL}.

\bibitem[{Touvron et~al.(2023)Touvron, Lavril, Izacard, Martinet, Lachaux,
  Lacroix, Rozi{\`e}re, Goyal, Hambro, Azhar et~al.}]{llama}
Hugo Touvron, Thibaut Lavril, Gautier Izacard, Xavier Martinet, Marie-Anne
  Lachaux, Timoth{\'e}e Lacroix, Baptiste Rozi{\`e}re, Naman Goyal, Eric
  Hambro, Faisal Azhar, et~al. 2023.
\newblock Llama: Open and efficient foundation language models.
\newblock \emph{arXiv preprint arXiv:2302.13971}.

\bibitem[{Wang et~al.(2022{\natexlab{a}})Wang, Cai, Jiang, Xie, Tu, and
  Lu}]{more}
Xinyu Wang, Jiong Cai, Yong Jiang, Pengjun Xie, Kewei Tu, and Wei Lu.
  2022{\natexlab{a}}.
\newblock Named entity and relation extraction with multi-modal retrieval.
\newblock In \emph{EMNLP}.

\bibitem[{Wang et~al.(2021{\natexlab{a}})Wang, Gui, Jiang, Jia, Bach, Wang,
  Huang, Huang, and Tu}]{ita}
Xinyu Wang, Min Gui, Yong Jiang, Zixia Jia, Nguyen Bach, Tao Wang, Zhongqiang
  Huang, Fei Huang, and Kewei Tu. 2021{\natexlab{a}}.
\newblock Ita: Image-text alignments for multi-modal named entity recognition.
\newblock \emph{arXiv preprint arXiv:2112.06482}.

\bibitem[{Wang et~al.(2021{\natexlab{b}})Wang, Jiang, Bach, Wang, Huang, Huang,
  and Tu}]{wang2021improving}
Xinyu Wang, Yong Jiang, Nguyen Bach, Tao Wang, Zhongqiang Huang, Fei Huang, and
  Kewei Tu. 2021{\natexlab{b}}.
\newblock Improving named entity recognition by external context retrieving and
  cooperative learning.
\newblock \emph{arXiv preprint arXiv:2105.03654}.

\bibitem[{Wang et~al.(2022{\natexlab{b}})Wang, Tian, Gui, Li, Ye, Yan, and
  Xiao}]{promptmner}
Xuwu Wang, Junfeng Tian, Min Gui, Zhixu Li, Jiabo Ye, Ming Yan, and Yanghua
  Xiao. 2022{\natexlab{b}}.
\newblock Prompt-based entity-related visual clue extraction and integration
  for multimodal named entity recognition.
\newblock In \emph{International Conference on Database Systems for Advanced
  Applications}, pages 297--305.

\bibitem[{Wang et~al.(2022{\natexlab{c}})Wang, Ye, Li, Tian, Jiang, Yan, Zhang,
  and Xiao}]{cat}
Xuwu Wang, Jiabo Ye, Zhixu Li, Junfeng Tian, Yong Jiang, Ming Yan, Ji~Zhang,
  and Yanghua Xiao. 2022{\natexlab{c}}.
\newblock Cat-mner: Multimodal named entity recognition with knowledge-refined
  cross-modal attention.
\newblock In \emph{ICME}, pages 1--6.

\bibitem[{Wei et~al.(2022)Wei, Wang, Schuurmans, Bosma, hsin Chi, Le, and
  Zhou}]{Wei2022ChainOT}
Jason Wei, Xuezhi Wang, Dale Schuurmans, Maarten Bosma, Ed~Huai hsin Chi, Quoc
  Le, and Denny Zhou. 2022.
\newblock Chain of thought prompting elicits reasoning in large language
  models.
\newblock \emph{ArXiv}, abs/2201.11903.

\bibitem[{Wu et~al.(2022)Wu, Xie, Zhou, Zhang, Chunping, Xu, and
  Zhang}]{metaaug4ner}
Linzhi Wu, Pengjun Xie, Jie Zhou, Meishan Zhang, Ma~Chunping, Guangwei Xu, and
  Min Zhang. 2022.
\newblock Robust self-augmentation for named entity recognition with meta
  reweighting.
\newblock In \emph{NAACL}, pages 4049--4060.

\bibitem[{Xu et~al.(2022)Xu, Huang, Sha, and Wang}]{maf}
Bo~Xu, Shizhou Huang, Chaofeng Sha, and Hongya Wang. 2022.
\newblock Maf: A general matching and alignment framework for multimodal named
  entity recognition.
\newblock In \emph{WSDM}, pages 1215--1223.

\bibitem[{Yao et~al.(2023)Yao, Yu, Zhao, Shafran, Griffiths, Cao, and
  Narasimhan}]{tot}
Shunyu Yao, Dian Yu, Jeffrey Zhao, Izhak Shafran, Thomas~L Griffiths, Yuan Cao,
  and Karthik Narasimhan. 2023.
\newblock Tree of thoughts: Deliberate problem solving with large language
  models.
\newblock \emph{arXiv preprint arXiv:2305.10601}.

\bibitem[{Yu et~al.(2020)Yu, Jiang, Yang, and Xia}]{twitter17}
Jianfei Yu, Jing Jiang, Li~Yang, and Rui Xia. 2020.
\newblock Improving multimodal named entity recognition via entity span
  detection with unified multimodal transformer.
\newblock ACL.

\bibitem[{Zhang et~al.(2021)Zhang, Li, Hu, Yang, Zhang, Wang, Choi, and
  Gao}]{vinvl}
Pengchuan Zhang, Xiujun Li, Xiaowei Hu, Jianwei Yang, Lei Zhang, Lijuan Wang,
  Yejin Choi, and Jianfeng Gao. 2021.
\newblock Vinvl: Revisiting visual representations in vision-language models.
\newblock In \emph{Proceedings of the IEEE/CVF Conference on Computer Vision
  and Pattern Recognition}, pages 5579--5588.

\bibitem[{Zhang et~al.(2018)Zhang, Fu, Liu, and Huang}]{twitter15}
Qi~Zhang, Jinlan Fu, Xiaoyu Liu, and Xuanjing Huang. 2018.
\newblock Adaptive co-attention network for named entity recognition in tweets.
\newblock In \emph{AAAI}, volume~32.

\bibitem[{Zhao et~al.(2022)Zhao, Li, Wu, Xing, and Dai}]{gcn}
Fei Zhao, Chunhui Li, Zhen Wu, Shangyu Xing, and Xinyu Dai. 2022.
\newblock Learning from different text-image pairs: A relation-enhanced graph
  convolutional network for multimodal ner.
\newblock In \emph{MM}, pages 3983--3992.

\bibitem[{Zheng et~al.(2021{\natexlab{a}})Zheng, Feng, Fu, Cai, Li, and
  Wang}]{Zheng2021MultimodalRE}
Changmeng Zheng, Junhao Feng, Ze~Fu, Yiru Cai, Qing Li, and Tao Wang.
  2021{\natexlab{a}}.
\newblock Multimodal relation extraction with efficient graph alignment.
\newblock \emph{MM}.

\bibitem[{Zheng et~al.(2021{\natexlab{b}})Zheng, Wu, Feng, Fu, and Cai}]{mnre}
Changmeng Zheng, Zhiwei Wu, Junhao Feng, Ze~Fu, and Yi~Cai. 2021{\natexlab{b}}.
\newblock Mnre: A challenge multimodal dataset for neural relation extraction
  with visual evidence in social media posts.
\newblock In \emph{ICME}, pages 1--6.

\bibitem[{Zhou et~al.(2022{\natexlab{a}})Zhou, Yang, Loy, and Liu}]{cocoop}
Kaiyang Zhou, Jingkang Yang, Chen~Change Loy, and Ziwei Liu.
  2022{\natexlab{a}}.
\newblock Conditional prompt learning for vision-language models.
\newblock In \emph{CVPR}, pages 16816--16825.

\bibitem[{Zhou et~al.(2022{\natexlab{b}})Zhou, Yang, Loy, and Liu}]{coop}
Kaiyang Zhou, Jingkang Yang, Chen~Change Loy, and Ziwei Liu.
  2022{\natexlab{b}}.
\newblock Learning to prompt for vision-language models.
\newblock \emph{IJCV}, 130(9):2337--2348.

\end{thebibliography}
\bibliographystyle{acl_natbib}

\end{document}